\documentclass[10pt,twocolumn,letterpaper]{article}

\usepackage{iccv}
\usepackage{times}
\usepackage{epsfig}
\usepackage{graphicx}
\usepackage{amsmath}
\usepackage{amssymb}

\usepackage{overpic}
\usepackage{caption}
\usepackage{csquotes}
\usepackage{rotating}
\usepackage{multirow}
\usepackage{booktabs}

\def\ie{\emph{i.e.}}
\def\eg{\emph{e.g.}}
\def\aka{\emph{a.k.a.}}

\newcommand{\red}[1]{{\textcolor{red}{#1}}}

\makeatletter
\def\cpartline#1{\@cpartline#1\@nil}
\def\@cpartline#1-#2\@nil{%
  \omit
  \@multicnt#1%
  \advance\@multispan\m@ne
  \ifnum\@multicnt=\@ne\@firstofone{&\omit}\fi
  \@multicnt#2%
  \advance\@multicnt-#1%
  \advance\@multispan\@ne
  \kern4pt
        \leaders\hrule\@height\arrayrulewidth\hfill
        \leaders\hrule\@height\arrayrulewidth\hfill
        \leaders\hrule\@height\arrayrulewidth\hfill
  \kern4pt
  \cr
  \noalign{\vskip-\arrayrulewidth}}
\makeatother

\usepackage[pagebackref=true,breaklinks=true,letterpaper=true,colorlinks,citecolor=ForestGreen,bookmarks=false]{hyperref}

\usepackage[dvipsnames]{xcolor} 

\def\ourTask{{{UCOS-DA}}}
\def\ourMod{{{FBA}}}

\iccvfinalcopy 

\def\iccvPaperID{8} 

\ificcvfinal\fi

\begin{document}

\title{Unsupervised Camouflaged Object Segmentation as Domain Adaptation}

\author{Yi Zhang\\
LIVIA, École de Technologie Supérieure\\
Montreal, Canada\\
\and
Chengyi Wu\\
Henan Polytechnic University\\
Henan, China\\
}


\maketitle
\ificcvfinal\thispagestyle{empty}\fi

\begin{abstract}
Deep learning for unsupervised image segmentation remains challenging due to the absence of human labels. The common idea is to train a segmentation head, with the supervision of pixel-wise pseudo-labels generated based on the representation of self-supervised backbones. By doing so, the model performance depends much on the distance between the distribution of target datasets, and the one of backbones' pre-training dataset (\eg, ImageNet).
In this work, we investigate a new task, namely unsupervised camouflaged object segmentation (UCOS), where the target objects own a common rarely-seen attribute, \ie, camouflage. Unsurprisingly, we find that the state-of-the-art unsupervised models struggle in adapting UCOS, due to the domain gap between the properties of generic and camouflaged objects.   
To this end, we formulate the \textbf{UCOS} as a source-free unsupervised \textbf{d}omain \textbf{a}daptation task (\textbf{\ourTask}), where both source labels and target labels are absent during the whole model training process. Specifically, we define a source model consisting of self-supervised vision transformers pre-trained on ImageNet. On the other hand, the target domain includes a simple linear layer (\ie, our target model) and unlabeled camouflaged objects. We then 
design a pipeline for foreground-background-contrastive self-adversarial domain adaptation, to achieve robust UCOS.
As a result, our baseline model achieves superior segmentation performance when compared with competing unsupervised models on the UCOS benchmark, with the training set which’s scale is only one tenth of the supervised COS counterpart. 
The UCOS benchmark and our baseline model are now publicly available\footnote{\url{https://github.com/Jun-Pu/UCOS-DA}}.
\end{abstract}

\section{Introduction}\label{sec:intro}

\begin{figure}[t!]
	\centering
	\begin{overpic}[width=0.48\textwidth]{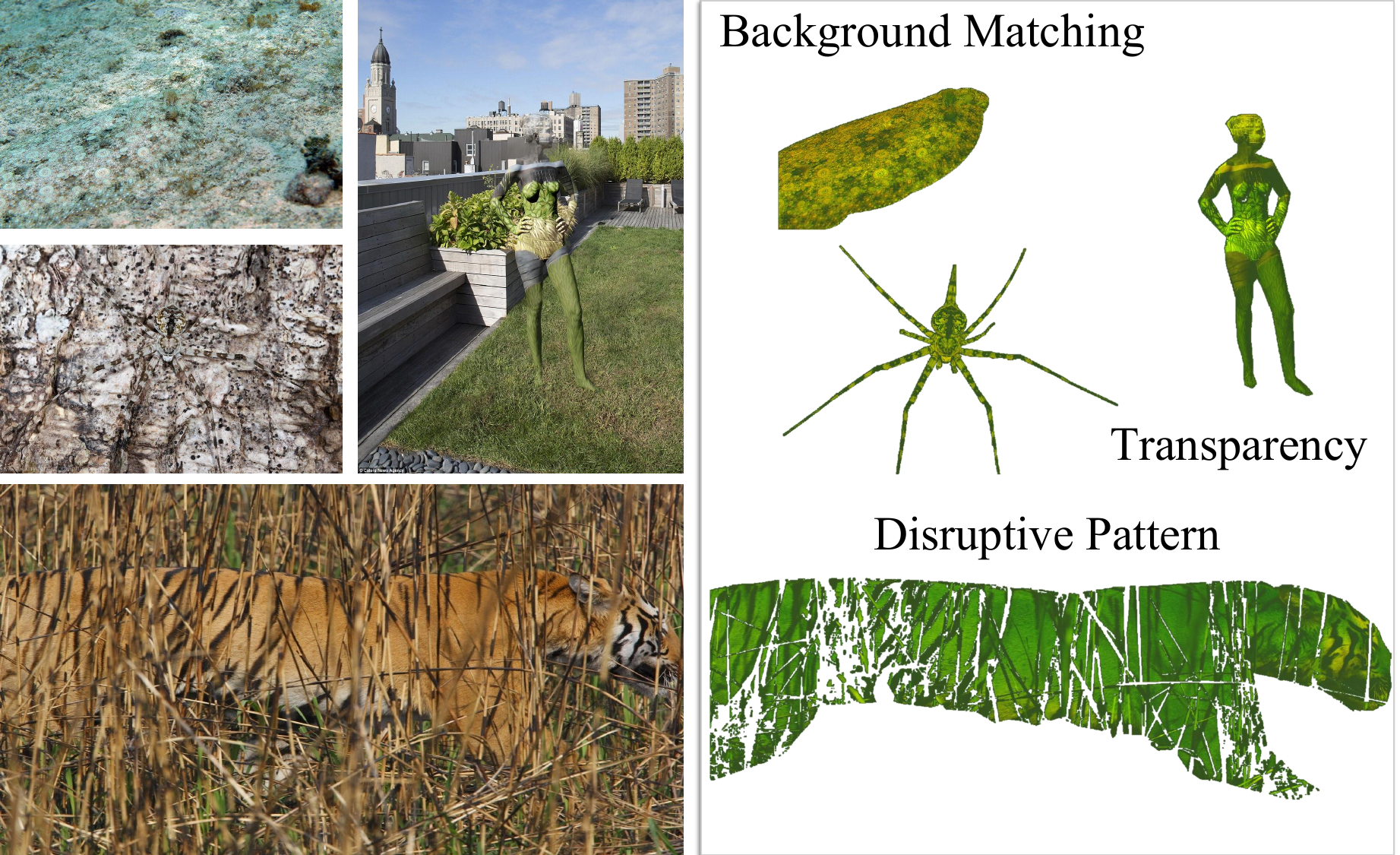}
    \end{overpic}
	\caption{An illustration of camouflaged object segmentation. The camouflage domain-specific properties (\eg, color-/texture-based background matching, transparency and disruptive pattern) are rarely-seen in generic object dataset such as ImageNet.}
    \label{fig:camo_attribute}
\end{figure}

In real-world scenes, there is a specific domain of objects which share one common attribute, namely \enquote{visual camouflage}. Camouflaged objects introduce challenges to image segmentation with their different types of concealing coloration \cite{cott1940adaptive} (Figure \ref{fig:camo_attribute}).
%
The common setting for camouflaged object segmentation (COS) is to fine-tune an encoder-decoder framework with well-labelled camouflaged objects \cite{fan2020camouflaged,wang2021d,sun2021c2fnet,zhang2023predictive}, based on the supervised ImageNet pre-trains \cite{imagenet_cvpr09,he2016deep,dosovitskiy2020image}. Though improvement \cite{zhang2023predictive,mao2021transformer} has been made as the booming development of vision transformers \cite{dosovitskiy2020image,ranftl2021vision}, this setting requires either dense labels (\ie, pixel-wise binary masks) or weak labels (\eg, points, object categories) as the supervision for training COS models. To advance COS to open-world applications where extensive human labels are hardly gained, and supervised models tend to be poorly generalized \cite{caron2021emerging,liu2021self,shi2023robust}; We take advantage of self-supervised ImageNet-based pre-trains \cite{caron2021emerging} and propose the first unsupervised COS baseline model, which requires no any human labels in the whole training pipeline.

\begin{figure}[t!]
	\centering
	\begin{overpic}[width=0.48\textwidth]{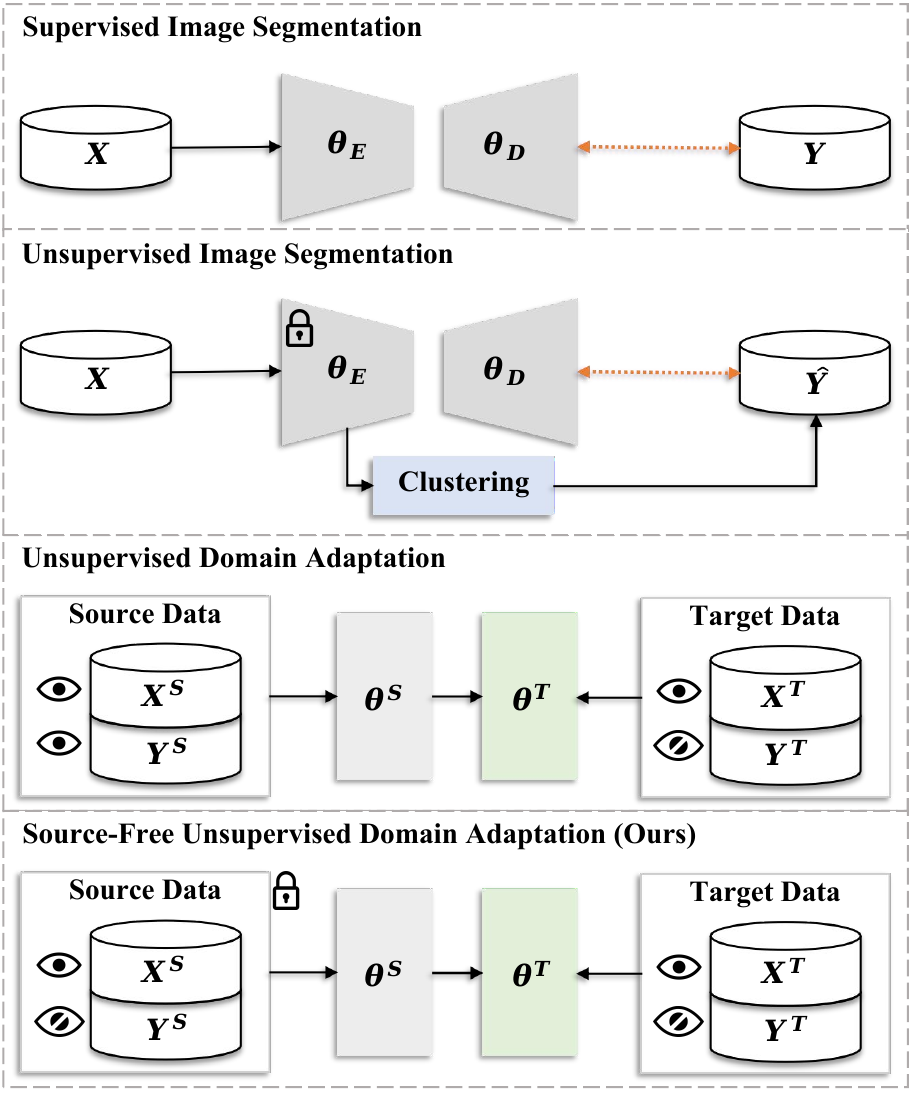}
    \end{overpic}
	\caption{An illustration of related tasks. $\{X,Y,\hat{Y}\}$ denote images, ground truth and pseudo-labels generated by unsupervised backbones, respectively. $\{\theta^{S},\theta^{T}\}$ indicate parameter sets of source and target models, respectively. $\{\theta_{E},\theta_{D}\}$ means the parameter set of encoder and decoder of a given segmentation network. Note that in our task, namely \ourTask, the source model ($\theta^{S}$) was trained in a self-supervised manner, without using source labels ($Y^{S}$).}
    \label{fig:task_illustration}
\end{figure}

Intuitively, we formulate the \textbf{u}nsupervised \textbf{COS} as a task of source-free unsupervised \textbf{d}omain \textbf{a}daptation, abbreviated as \textbf{\ourTask}~(Figure \ref{fig:task_illustration}).
Being different to common source-free unsupervised domain adaptation settings where human labels are needed to train the source model, our \ourTask~setting does not involve supervised training of the source domain. To conduct the new task, we propose a \ourTask~baseline model consisting of three components, \ie, a self-supervised source model, a light-weighted target model and an {a}dversarial {d}omain {a}daptation module (Figure \ref{fig:UCOSDA_pipeline}). Following state-of-the-art unsupervised image segmentation methods \cite{wang2022self,melas2022deep,henaff2022object,Shin_2022_CVPR,simeoni2022unsupervised,seitzer2022bridging,lv2023weakly}, we use DINO \cite{caron2021emerging}'s ImageNet pre-trained self-supervised vision transformer, as the unsupervised object-centric feature extractor (\ie, our source model). 
Considering the ambiguity (Figure \ref{fig:camo_attribute}) between object parts and background region in COS, we explore to shift more attention to the local features representing the boundary of camouflaged targets, during domain adaptation. We thus design a self-adversarial training module to weight more importance to the boundary-specific object-centric representations. 
Meanwhile, the target model learns to segment camouflaged objects with the pixel-wise supervision of pseudo-labels gained from DINO features.

In a nutshell, by proposing the new task (\ourTask) in the context of fully unsupervised image segmentation, we investigate the domain transfer ability of state-of-the-art self-supervised vision transformers, especially towards the circumstance where a large discrepancy exists between the source domain and target domain (here we mean different visual patterns between generic objects and camouflaged objects). The main contributions are summarized as follows:
\textbf{1)} We firstly investigate the task of unsupervised COS, by implementing a systematic benchmark study involving seven evaluative metrics and five state-of-the-art image segmentation methods.
\textbf{2)} We investigate unsupervised COS from a perspective of source-free unsupervised domain adaptation, by proposing a baseline model which gains competitive results on multiple benchmark datasets.
Besides, we discuss key issues for bridging domain adaptation to unsupervised object-centric representation learning. We hope our work could inspire more generalizable unsupervised image segmentation models in future researches.
%

\section{Related Work}\label{sec:related_work}

\subsection{Self-Supervised Representation Learning}\label{sec:SSL}
Learning to localize objects without using any human labels is a longstanding issue in the field of computer vision. The issue has recently appealed much more attention from the community, owing to the release of self-supervised representation learning methodologies, such as \enquote{MoCo Trilogy} \cite{he2020momentum,chen2020improved,chen2021empirical}, SimCLR \cite{chen2020simple}, DenseCL \cite{wang2021dense}, DINO \cite{caron2021emerging}, MAE \cite{he2022masked} and \enquote{BEiT Trilogy} \cite{bao2021beit,peng2022beit,wang2023image}. These models were trained with large-scale datasets (\eg, ImageNet \cite{imagenet_cvpr09}) in a self-supervised manner, advancing the label-free object discovery. We briefly summarize recent self-supervised methods according to their types of pretext tasks:

\noindent\textbf{Contrastive Learning.} 
Pioneer works, MoCo \cite{he2020momentum} and SimCLR \cite{chen2020simple}, proposed to optimize their networks' features via calculating similarities between two branches of features, respectively acquired from two sets of visual inputs.
Notably, MoCo \cite{he2020momentum} used two encoders with different parameter updating strategies
, while SimCLR \cite{chen2020simple} took advantage of one encoder with two sets of parameters (Siamese framework).
Following MoCo, DenseCL \cite{wang2021dense} proposed dense projection heads to facilitate downstream unsupervised dense prediction tasks.
Inspired by both MoCo and SimCLR, BYOL \cite{grill2020bootstrap} used an on-line network and a target network to conduct self-supervised training, without relying on negative pairs.
Following BYOL, DINO \cite{caron2021emerging} applied two interactive encoders sharing the same ViT \cite{dosovitskiy2020image}-based architecture however with different parameter sets and updating strategies, achieved representations that illustrate superior object emergence when compared to the fully supervised counterparts.

\noindent\textbf{Masked Image Modeling (MIM).}
MIM-based methods aim to learn representations via reconstructing original images from image patches where a certain percentage of them are masked out. BeiT \cite{bao2021beit}, as one of the pioneer works within this category, followed the masked language modeling strategy proposed in BERT \cite{devlin2018bert} and introduced MIM into vision transformers. MAE \cite{he2022masked} also proposed auto-encoder-like architecture but to reconstruct pixels rather than to predict tokens. BEiT-v2 \cite{peng2022beit} replaced the original reconstruction target with semantic-rich visual tokenizers to learn representations highlighting semantic cues. MaskFeat \cite{wei2022masked} also used MIM for model training however with the optimizing target of reconstructing HOG features of the masked image patches. SimMIM \cite{xie2022simmim} proposed new prediction head consisting of only one linear payer. 

\noindent\textbf{Multi-Modal Alignment.}
The community recently witnessed a competition in establishing large vision-language models (VLMs) for representation learning \cite{radford2021learning,xu2022groupvit,li2022grounded,tschannen2023clippo,dong2023maskclip,wang2023image,kim2023region}. Compared to vision-only self-supervised settings, VLMs relax the constrict of leveraging human labels by relying on image-text pairs, to learn multi-modal representations via aligning visual and textual cues.  
CLIP \cite{radford2021learning} jointly trained a text encoder and an image encoder to predict positive image-text pairs, achieving state-of-the-art zero-shot image classification. To further obtain object-centric locality-aware representation, GLIP \cite{li2022grounded} jointly optimized image and text encoders to localize positive region-word pairs. GroupViT \cite{xu2022groupvit} added grouping blocks to each level of a ViT \cite{dosovitskiy2020image}, enabling progressive optimization of its vision encoder with only text-based weak supervision. Being different to above frameworks which rely on separate text and image encoders, CLIPPO \cite{tschannen2023clippo} extracted both image and text features with a single encoder. Methods such as MaskCLIP \cite{dong2023maskclip} and BEiT-v3 \cite{wang2023image} combined MIM strategy and visual-language contrastive learning to pursue generalizable representation. RO-ViT \cite{kim2023region} achieved state-of-the-art open-vocabulary object detection, via manipulating ViT's positional embeddings at the pre-training stage and gaining region-aware image-text pairs at the fine-tuning stage. More recently, MUG \cite{zhao2023vision} achieved new state-of-the-art in vision transfer learning tasks, via training a self-supervised vision-language model based on large-scale web data.

Despite the booming development of large-scale self-supervised multi-modal pre-trained models, unsupervised domain adaptation remains an open issue due to the finite scale of the pre-training data. To this end, OOD-CV \cite{zhao2022ood} released an open challenge\footnote{\url{http://www.ood-cv.org/challenge.html}} to continually advance researches in exploring the transfer learning ability of state-of-the-art self-supervised pre-trained models.  

\subsection{State-of-the-Art Unsupervised Segmentation}\label{sec:SOTA_UnSSeg}
The \enquote{pre-training and fine-tuning} has been the most commonly-used paradigm for training deep neural networks since the emergence of ImageNet \cite{imagenet_cvpr09}. Recent development of self-supervised pre-trains (Section \ref{sec:SSL}) stimulates the development of unsupervised image segmentation \cite{STEGO,wang2022self,melas2022deep,wang2022freesolo,yun2022patch,ziegler2022self,yin2021transfgu,Shin_2022_CVPR,henaff2022object,seitzer2022bridging,simeoni2022unsupervised,Ishtiak_2023_CVPR,wang2023cut}. These methods are able to conduct instance-level pixel-wise classification without using any manual annotations.

\noindent\textbf{Unsupervised Object Segmentation.} 
TokenCut \cite{wang2022self} conducted spectral clustering based on the DINO \cite{caron2021emerging} features, yet the method is able to segment only one object per image. SelfMask \cite{Shin_2022_CVPR} applied different number of clusters to produce multiple binary masks, and introduced a voting strategy to gain the final prediction. Also based on DINO features, FOUND \cite{simeoni2022unsupervised} retrieved the background seed and identified its complement as the foreground. Final results were obtained by training a linear layer with the supervision of retrieved foreground masks. DINOSAUR \cite{seitzer2022bridging} explored the task from a perspective of object-centric learning. The method was optimized to reconstruct the given images with slot-attention-based \cite{locatello2020object} decomposed object-centric representations. There is another class of methods \cite{voynov2021object,abdal2021labels4free,he2022ganseg,bae2022furrygan,zou2022ilsgan} that use generative adversarial networks to generate the foreground masks representing target objects. Though progress was achieved during the past few years, we find that current unsupervised object segmentation methods tend to fail the cases where objects show complicated appearances in specific context (\eg, camouflage, an object-centric attribute rarely-seen in ImageNet).  

\noindent\textbf{Unsupervised Semantic Segmentation.}
Thanks to the booming trend of large-scale self-supervised pre-trained models, the community witnessed an important change of the learning paradigm of semantic segmentation, from fully-/weakly-supervised learning to fully unsupervised learning. Recent methods such as STEGO \cite{STEGO}, SpectralSeg \cite{melas2022deep}, FreeSOLO \cite{wang2022freesolo}, SelfPatch \cite{yun2022patch}, TransFGU \cite{yin2021transfgu}, Leopart \cite{ziegler2022self}, Odin \cite{henaff2022object}, Exemplar-FreeSOLO \cite{Ishtiak_2023_CVPR} and CutLER \cite{wang2023cut}, were trained to assign each pixel to specific object class without the supervision of any human labels. Similar to supervised methods, current unsupervised semantic segmentation methods face challenges such as occlusion detection, small object detection and multi-instance identification.

\subsection{Unsupervised Domain Adaptation}\label{sec:SOTA_SFUDA}
Recent researches \cite{kundu2020universal,yang2021generalized,xia2021adaptive,yang2021exploiting,qiu2021source,lo2023spatio,wang2023mhpl,karim2023c,shen2023balancing} investigated source-free unsupervised domain adaptation, where only the pre-trained source model and unlabeled target data are accessible during the domain adaptation. USFDA \cite{kundu2020universal} proposed a source similarity metric to conduct domain adaptation without source data, and achieved on-par results when compared to the source-dependent counterparts. G-SFDA \cite{yang2021generalized} proposed local structure clustering to adapt source model to the target domain in the absence of source data. A$^{2}$Net \cite{xia2021adaptive} was trained to classify the target data into source-similar and source-dissimilar groups, via an adaptive adversarial strategy. NRC-SFDA \cite{yang2021exploiting} explored the local affinity of target data and achieved improved source-free adaptation upon both 2D and 3D target data. CPGA \cite{qiu2021source} disentangled the source model and gained class-wise features, namely avatar prototype, to facilitate source-target alignment. More recently, STPL \cite{lo2023spatio} used temporal cues, \ie, optical flow, to conduct domain adaptation for video semantic segmentation. ASFDA \cite{wang2023mhpl} resorted to active learning technique to identify a small set of source features, which supported the efficient training of the target model. C-SFDA \cite{karim2023c} proposed new self-training strategy based on curriculum learning. MSFDA \cite{shen2023balancing} explored multi-source-free domain adaptation and found an inherent bias-variance trade-off within the task, thus inspiring future works.

\subsection{Uniqueness of Our Model}\label{sec:unique_ours}
Training a segmentation head, with merely unlabeled COS dataset and ImageNet pre-trained self-supervised model, can be regarded as a source-free unsupervised domain adaptation task. Due to the out-of-distribution properties of camouflaged objects (Figure \ref{fig:camo_attribute}), unsupervised COS is an extremely challenging task. To this end, we consider 
to discovery and reserve the boundary-specific local self-supervised features, 
and resort to adversarial domain adaptation technique to improve the model transfer robustness. Besides the innovation towards the task formulation and camouflaged prior modeling, we define our target model as a simple linear layer yet predicts superior results when compared with its counterparts in unsupervised object segmentation.

\section{UCOS-DA Methodology}

\begin{figure}[t!]
	\centering
	\begin{overpic}[width=0.48\textwidth]{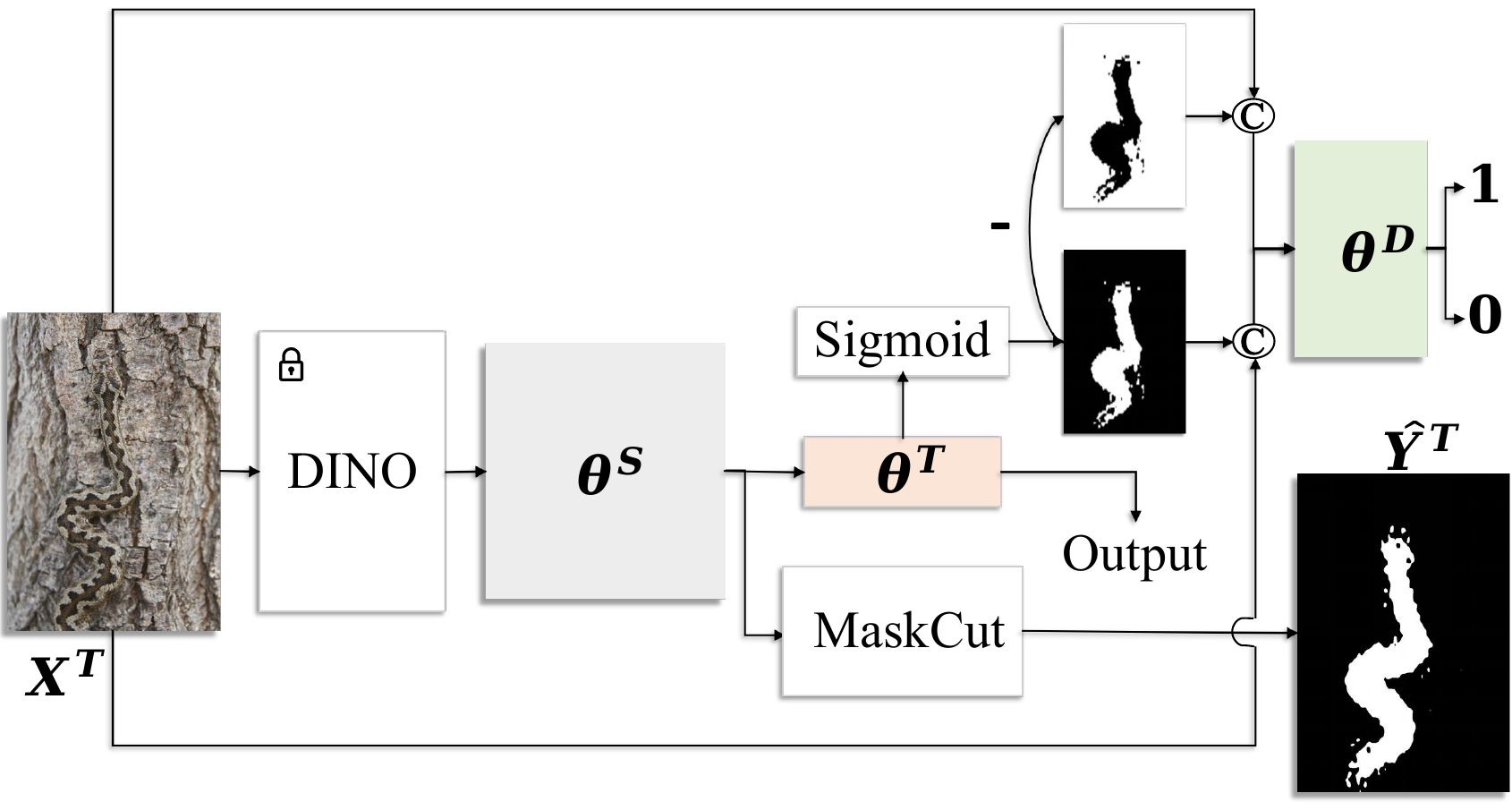}
    \end{overpic}
	\caption{The pipeline of our proposed \ourTask~baseline model. The model consists of a frozen source model ($\theta^{S}$), a light-weighted linear target model ($\theta^{T}$) and a {f}oreground-{b}ackground-contrastive self-adversarial domain {a}daptation module ($\theta^{D}$). Notably, no any human labels are used for \ourTask~pseudo-labelling, pre-training or fine-tuning.}
    \label{fig:UCOSDA_pipeline}
\end{figure}

We propose the first baseline model for \textbf{u}nsupervised \textbf{c}amouflaged \textbf{o}bject \textbf{s}egmentation, from the perspective of \textbf{d}omain \textbf{a}daptation (\textbf{\ourTask}). The model consists of a self-supervised ImageNet-based pre-trained vision transformer as the source model ($\theta^{S}$), a linear-probe layer as the target model ($\theta^{T}$), and a \textbf{f}oreground-\textbf{b}ackground-contrastive self-adversarial domain \textbf{a}daptation module ($\theta^{D}$, abbreviated as \textbf{\ourMod}). The pipeline of the proposed baseline model is shown in Figure \ref{fig:UCOSDA_pipeline}.

\subsection{UCOS-DA Motivation\&Formulation}
A popular chatbot gives a definition towards \enquote{object}: \enquote{\emph{An object refers to a distinct item or entity that occupies space, has properties, can be perceived through our senses}}. In 2D domain, object segmentation (\aka, object-level pixel-wise classification) models usually require manual annotations as supervision to learn the mapping from images to objects. As the recent development of self-supervised models, it is inspiring to see that, specific pretext tasks (\eg, enforcing the view-invariance \cite{grill2020bootstrap,caron2021emerging}, recovering the missing parts \cite{he2022masked}), enable deep learning models to discover object concepts without using external supervision of human labels. Thus, self-supervised learning seems to be a more humanoid learning paradigm and thus promising.
In the context of unsupervised COS, we aim to achieve a model which learns the camouflaged properties with only unlabelled image data, thus segmenting objects concealed in various real-world scenes effectively. Considering the absence of large-scale camouflage pre-trains, a feasible solution is to extract features from generic data-based self-supervised models and adapt them to the camouflage domain. 

To this end, we formulate the objective of \ourTask~as minimizing an empirical loss function:
\begin{equation}
\label{ucosda_objective}
\begin{aligned}
\min_{\{\theta^{S},\theta^D,\theta^{T}\}}\mathbb{E}_{X^T,\hat{Y}^T}[\mathcal{L}(f^{T}(X^T;\theta^S,\theta^D,\theta^T),\hat{Y}^T)]~~~~~~~~~~~~\\
=\int\mathcal{L}(f^T(X^T;\theta^S,\theta^D,\theta^T), \hat{Y}^T)d p(X^T,\hat{Y}^T)~~~~~~~~~\\
\approx\frac{1}{N}\sum_{i=1}^N\mathcal{L}(f^T(x_i^{T};\theta^S,\theta^D,\theta^T),\hat{y_i}^T),~~~~~~~~~~~~~~~~~~~~~~
\end{aligned}
\end{equation}
with
\begin{equation}
\label{ucosda_objective_supp}
(x_i^T, \hat{y_i}^T) \sim p(X^T,\hat{Y}^T),
\end{equation}
where $\{\theta^S,\theta^D,\theta^T\}$ denotes parameter sets of the source model, the \ourMod~module and the target model, respectively. $(x_i^T, \hat{y}_i^T)$ denotes a sample pair from the joint data distribution in the target domain. Notably, the $\hat{Y}^T$ indicates the pseudo-labels corresponding to the training data in the target domain. $\mathcal{L}(\cdot)$ means the loss function.

\subsection{UCOS-DA Architecture}

\begin{figure}[t!]
	\centering
	\begin{overpic}[width=0.47\textwidth]{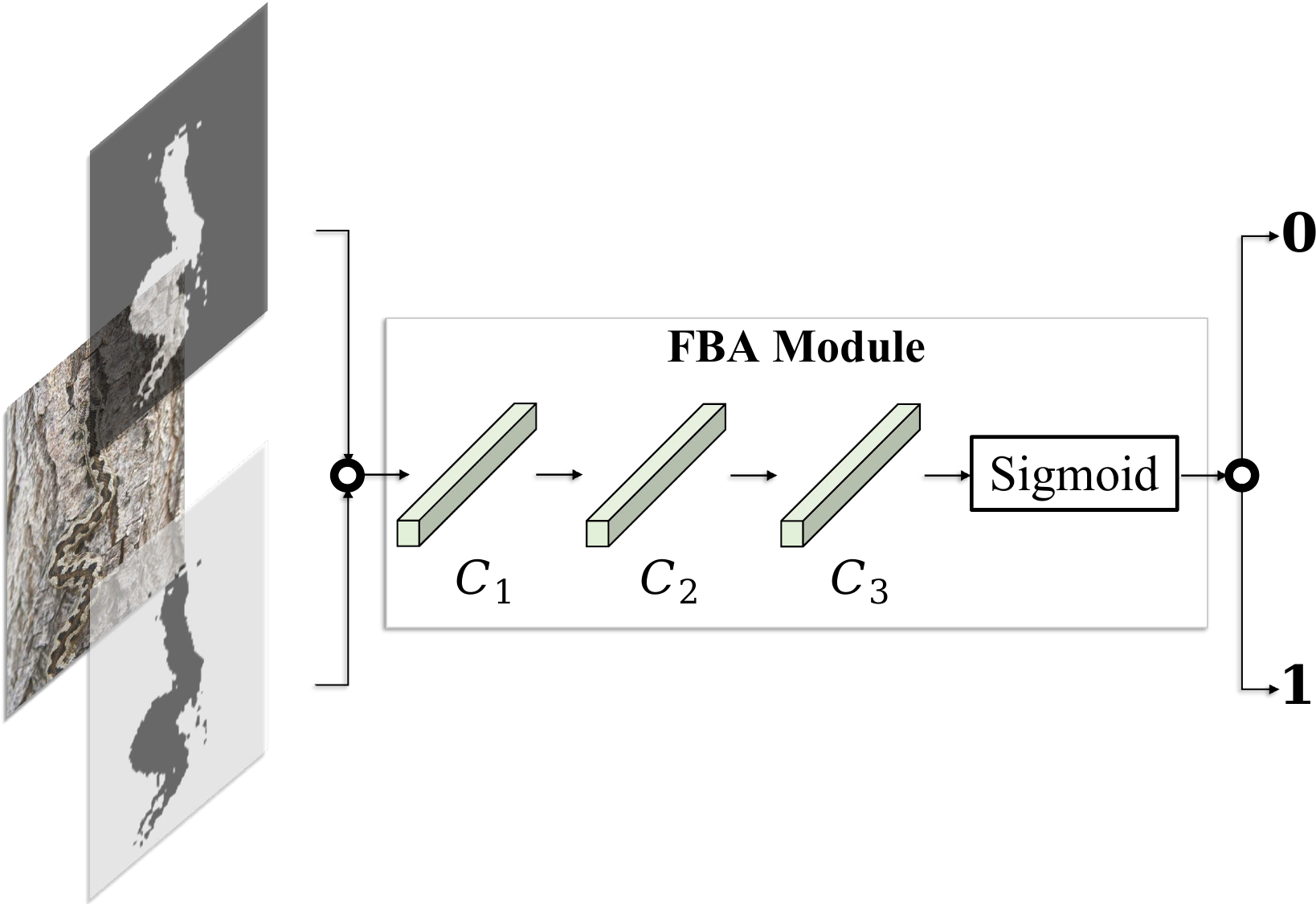}
    \end{overpic}
	\caption{The architecture of the \ourMod~(\textbf{f}oreground-\textbf{b}ackground-contrastive self-adversarial domain \textbf{a}daptation) module ($\theta^{D}$). $\{C_1, C_2, C_3\}$ denotes the number of channels of each linear layer, respectively.}
    \label{fig:FBA_module}
\end{figure}

\noindent\textbf{Generic Object-Centric Knowledge Extraction.}
Acorrding to previous researches \cite{wang2022self,simeoni2022unsupervised,wang2023cut}, among various self-supervised pre-trains, DINO \cite{caron2021emerging} has proved its superior object emergence ability and is regarded as one of most promising candidates for downstream unsupervised image segmentation tasks. We use DINO ImageNet pre-trains as our source model, and extract its rich generic object knowledge to generate pseudo-labels, and to facilitate a self-supervised training of the target model.

\noindent\textbf{Pseudo-Labels.}
We resort to normalized cuts technique \cite{shi2000normalized} to generate coarse maps based on DINO features. Specifically, we resort to MaskCut methodology \cite{wang2023cut}, which conducts multiple iterations of normalized cuts with DINO features, based on a patch-level affinity matrix.


\noindent\textbf{Adversarial Domain Adaptation.}
To adapt DINO pre-trained features to unsupervised COS, we first study the object priors when it comes to the camouflaged scenario. In fact, animals tend to deceive predators' visual perception with specific concealing coloration. As a consequence, noisy visual cues are brought to the boundary region of camouflaged objects in 2D images, making it hard to obtain on-par segmentation results. We argue that the blur of object boundary is one of the main cause for the 
big divergence between camouflaged and generic data distributions. 

To close the gap between the source (generic) domain and the target (camouflage) domain, we suggest a new module to emphasize the reservation of boundary-specific local representations of source model, during training the target model. Specifically, we introduce a foreground-background-contrastive self-adversarial domain adaptation (FBA) module (Figure \ref{fig:FBA_module}), to conduct a sub-task aiming at further distinguishing the predicted foreground maps from their complements.
Our FBA module mainly consists of three hierarchical linear layers, computing foreground/background class score (${S}, {S} \in [0,1]$) as:
\begin{equation}
\label{fba_formula}
{S} = \sigma(FC_{C3}(LR(FC_{C2}(LR(FC_{C1}(Cat(X,P'))))))),
\end{equation}
where $\{X,P'\}$ means the given images and corresponding binary masks gained via the target model. $\sigma(\cdot)$, $FC(\cdot)$, $LR(\cdot)$ and $Cat(\cdot)$ denotes Sigmoid function, linear(fully-connected) layer, leakyReLu activation layer and concatenation operation, respectively.

\subsection{Implementation Details}

\noindent\textbf{Loss Function.} As the target model and \ourMod~module are co-trained for domain adaptation, the total loss ($\mathcal{L}$) of our \ourTask~baseline model is thus formulated as the sum of a segmenting loss ($\mathcal{L}^{Seg.}$) and an adversarial loss ($\mathcal{L}^{Adv.}$):
\begin{equation}
\label{loss_func}
\mathcal{L} = \mathcal{L}^{Seg.}(P, \hat{Y}^T) + \mathcal{L}^{Adv.}(S, C),
\end{equation}
where $P$ and $C$ ($C \in \{0,1\}$) denotes segmentation results of target model, and foreground/background class label, respectively. Notably, in this work, we apply the structure loss \cite{wei2020f3net} as the segmentation loss $\mathcal{L}^{Seg.}$, and binary cross entropy loss as the foreground/background classification loss (\ie, adversarial loss $\mathcal{L}^{Adv.}$).

\noindent\textbf{Hyper-Parameters.} We train the \ourTask~baseline model by using PyTorch with a maximum epoch of 5. The images are re-scaled to the size of 512$\times$512 during training. The initial learning rate of the target model and the FBA module is set to 5e-3 and 5e-4, respectively.

\section{Experiments}

\begin{table*}[t!]
  \centering
 \setlength\tabcolsep{3pt}
 \caption{
   Comparison of our \ourTask~and state-of-the-art unsupervised methods on salient object segmentation benchmark datasets. The \textbf{best} and the \underline{second best} results of each row are highlighted. 
   }
   \label{tab:benchmark_UCOSDA_SOD}
  \resizebox{1\textwidth}{!}{
  \begin{tabular}{ccrcccccccc}
   \toprule
    \multirow{2}{*}{Task} & \multirow{2}{*}{Dataset} & \multirow{2}{*}{Metric}
   & BigGW 
   & TokenCut 
   & TokenCut w/ B.S. 
   & SpectralSeg 
   & SelfMask
   & SelfMask w/ U.B. 
   & FOUND
   & \ourTask(\textbf{Ours})
   \\
   &&& ICML'21 \cite{voynov2021object} & CVPR'22 \cite{wang2022self} & CVPR'22 \cite{wang2022self} & CVPR'22  \cite{melas2022deep} & CVPRw'22 \cite{Shin_2022_CVPR} & CVPRw'22 \cite{Shin_2022_CVPR} & CVPR'23 \cite{simeoni2022unsupervised} & ICCVw'23
   \\
   
  \hline
  \multirow{18}{*}{\begin{sideways}Salient Object Segmentation\end{sideways}}
  
  & \multirow{9}{*}{\begin{sideways}ECSSD \cite{shi2015hierarchical}\end{sideways}}
  & mIoU~$\uparrow$ & .689 & .712 & .774 & .733 & .779 & .787 & \underline{.805} & \textbf{.816}
  \\
  && Acc.~$\uparrow$ & .905 & .918 & .934 & .891 & .943 & .946 & \underline{.948} & \textbf{.951}
  \\
  && $F_{\beta}^{max}~\uparrow$ & .800 & .803 & .874 & .805 & .892 & \textbf{.897} & \underline{.896} & .891
  \\
  && $F_{\beta}^{mean}~\uparrow$ & .654 & .801 & .714 & .803 & .861 & .867 & \textbf{.894} & \underline{.888}
  \\
  && $F_{\beta}^{W}~\uparrow$ & .568 & .785 & .630 & .790 & .846 & .852 & \textbf{.877} & \underline{.876}
  \\
  && $S_{\alpha}~\uparrow$ & .783 & .807 & .832 & .806 & .866 & .871 & \underline{.875} & \textbf{.878}
  \\
  && $E_{\phi}^{max}~\uparrow$ & .871 & .886 & .905 & .865 & .928 & \underline{.932} & \underline{.932} & \textbf{.934}
  \\
  && $E_{\phi}^{mean}~\uparrow$ & .714 & .884 & .755 & .862 & .920 & .925 & \underline{.930} & \textbf{.931}
  \\
  && $\mathcal{M}~\downarrow$ & .169 & .082 & .129 & .109 & .058 & .055 & \underline{.052} & \textbf{.049}
  \\

  \cpartline{3-11}
  & \multirow{9}{*}{\begin{sideways}HKU-IS \cite{li2015visual}\end{sideways}}
  & mIoU~$\uparrow$ & .641 & .608 & .673 & .735 & .747 & .755 & \underline{.787} & \textbf{.794}
  \\
  && Acc.~$\uparrow$ & .905 & .916 & .936 & .932 & .949 & .951 & \underline{.958} & \textbf{.959}
  \\
  && $F_{\beta}^{max}~\uparrow$ & .760 & .741 & .832 & .815 & .869 & \underline{.874} & \textbf{.877} & .872 
  \\
  && $F_{\beta}^{mean}~\uparrow$ & .611 & .739 & .667 & .812 & .830 & .836 & \textbf{.875} & \underline{.870}
  \\
  && $F_{\beta}^{W}~\uparrow$ & .515 & .703 & .557 & .801 & .818 & .824 & \textbf{.863} & \underline{.861}
  \\
  && $S_{\alpha}~\uparrow$ & .761 & .748 & .777 & .828 & .851 & .856 & \underline{.869} & \textbf{.871}
  \\
  && $E_{\phi}^{max}~\uparrow$ & .859 & .866 & .871 & .896 & .930 & .934 & \textbf{.939} & \underline{.937}
  \\
  && $E_{\phi}^{mean}~\uparrow$ & .696 & .864 & .728 & .894 & .919 & .923 & \textbf{.936} & \underline{.935}
  \\
  && $\mathcal{M}~\downarrow$ & .166 & .084 & .123 & .068 & .052 & .050 & \underline{.042} & \textbf{.041}
  \\
  
  \bottomrule
  \end{tabular}}
\end{table*}

\begin{table*}[t!]
  \centering
 \setlength\tabcolsep{3pt}
 \caption{
   Comparison of our \ourTask~baseline and state-of-the-art unsupervised methods on camouflaged object segmentation benchmark datasets. The \textbf{best} and the \underline{second best} results of each row are highlighted. 
   }
   \label{tab:benchmark_UCOSDA_COD}
  \resizebox{1\textwidth}{!}{
  \begin{tabular}{ccrcccccccc}
   \toprule
   \multirow{2}{*}{Task} & \multirow{2}{*}{Dataset} & \multirow{2}{*}{Metric}
   & BigGW 
   & TokenCut 
   & TokenCut w/ B.S. 
   & SpectralSeg 
   & SelfMask
   & SelfMask w/ U.B. 
   & FOUND
   & \ourTask(\textbf{Ours})
   \\
   &&& ICML'21 \cite{voynov2021object} & CVPR'22 \cite{wang2022self} & CVPR'22 \cite{wang2022self} & CVPR'22  \cite{melas2022deep} & CVPRw'22 \cite{Shin_2022_CVPR} & CVPRw'22 \cite{Shin_2022_CVPR} & CVPR'23 \cite{simeoni2022unsupervised} & ICCVw'23
   \\
  
   \hline
  \multirow{36}{*}{\begin{sideways}Camouflaged Object Segmentation\end{sideways}}
   
  & \multirow{9}{*}{\begin{sideways}CAMO \cite{le2019anabranch}\end{sideways}}
  & mIoU~$\uparrow$ & .322 & .431 & .422 & .411 & .418 & .430 & \underline{.505} & \textbf{.528}
  \\
  && Acc.~$\uparrow$ & .775 & .837 & .838 & .765 & .813 & .819 & \underline{.871} & \textbf{.873}
  \\
  && $F_{\beta}^{max}~\uparrow$ & .428 & .546 & .550 & .486 & .549 & .561 & \underline{.635} & \textbf{.647}
  \\
  && $F_{\beta}^{mean}~\uparrow$ & .349 & .543 & .434 & .481 & .536 & .547 & \underline{.633} & \textbf{.646}
  \\
  && $F_{\beta}^{W}~\uparrow$ & .299 & .498 & .383 & .450 & .483 & .495 & \underline{.584} & \textbf{.606}
  \\
  && $S_{\alpha}~\uparrow$ & .565 & .633 & .639 & .579 & .617 & .627 & \underline{.685} & \textbf{.701}
  \\
  && $E_{\phi}^{max}~\uparrow$ & .678 & .708 & .699 & .658 & .713 & .724 & \underline{.784} & \textbf{.786}
  \\
  && $E_{\phi}^{mean}~\uparrow$ & .528 & .706 & .595 & .648 & .698 & .708 & \underline{.782} & \textbf{.784}
  \\
  && $\mathcal{M}~\downarrow$ & .282 & .163 & .195 & .235 & .188 & .182 & \underline{.129} & \textbf{.127}
  \\
 
  \cpartline{3-11}
  & \multirow{9}{*}{\begin{sideways}CHAMELEON \cite{Chameleon2018}\end{sideways}}
  & mIoU~$\uparrow$ & .267 & .436 & .415 & .381 & .396 & .406 & \underline{.468} & \textbf{.525}
  \\
  && Acc.~$\uparrow$ & .807 & .868 & \underline{.871} & .780 & .825 & .832 & \textbf{.905} & \textbf{.905}
  \\
  && $F_{\beta}^{max}~\uparrow$ & .356 & .540 & .544 & .446 & .511 & .522 & \underline{.591} & \textbf{.631}
  \\
  && $F_{\beta}^{mean}~\uparrow$ & .294 & .536 & .393 & .440 & .481 & .491 & \underline{.590} & \textbf{.629}
  \\
  && $F_{\beta}^{W}~\uparrow$ & .244 & .496 & .351 & .410 & .436 & .447 & \underline{.542} & \textbf{.591}
  \\
  && $S_{\alpha}~\uparrow$ & .547 & .654 & .655 & .575 & .619 & .629 & \underline{.684} & \textbf{.715}
  \\
  && $E_{\phi}^{max}~\uparrow$ & .662 & .743 & .734 & .638 & .726 & .734 & \textbf{.812} & \underline{.804}
  \\
  && $E_{\phi}^{mean}~\uparrow$ & .527 & .740 & .582 & .628 & .675 & .683 & \textbf{.810} & \underline{.802}
  \\
  && $\mathcal{M}~\downarrow$ & .257 & \underline{.132} & .169 & .220 & .176 & .169 & \textbf{.095} & \textbf{.095}
  \\
 
  \cpartline{3-11}
  & \multirow{9}{*}{\begin{sideways}COD10K \cite{fan2020camouflaged}\end{sideways}}
  & mIoU~$\uparrow$ & .236 & .415 & .423 & .331 & .388 & .397 & \underline{.428} & \textbf{.462}
  \\
  && Acc.~$\uparrow$ & .798 & .897 & .903 & .807 & .870 & .875 & \textbf{.915} & \underline{.914}
  \\
  && $F_{\beta}^{max}~\uparrow$ & .315 & .509 & \underline{.537} & .395 & .504 & .514 & .521 & \textbf{.548}
  \\
  && $F_{\beta}^{mean}~\uparrow$ & .246 & .502 & .399 & .388 & .469 & .478 & \underline{.520} & \textbf{.546}
  \\
  && $F_{\beta}^{W}~\uparrow$ & .185 & .469 & .334 & .360 & .431 & .440 & \underline{.482} & \textbf{.513}
  \\
  && $S_{\alpha}~\uparrow$ & .528 & .658 & .666 & .575 & .637 & .645 & \underline{.670} & \textbf{.689}
  \\
  && $E_{\phi}^{max}~\uparrow$ & .670 & .740 & .739 & .606 & .718 & .728 & \textbf{.753} & \underline{.741}
  \\
  && $E_{\phi}^{mean}~\uparrow$ & .497 & .735 & .609 & .595 & .679 & .687 & \textbf{.751} & \underline{.740}
  \\
  && $\mathcal{M}~\downarrow$ & .261 & .103 & .127 & .193 & .131 & .125 & \textbf{.085} & \underline{.086}
  \\
 
  \cpartline{3-11}
  & \multirow{9}{*}{\begin{sideways}NC4K \cite{yunqiu2021ranking}\end{sideways}}
  & mIoU~$\uparrow$ & .382 & .546 & .561 & .495 & .529 & .538 & \underline{.566} & \textbf{.590}
  \\
  && Acc.~$\uparrow$ & .814 & .899 & .904 & .841 & .887 & .891 & \textbf{.916} & \underline{.915}
  \\
  && $F_{\beta}^{max}~\uparrow$ & .484 & .655 & .682 & .570 & .661 & .670 & \underline{.676} & \textbf{.691}
  \\
  && $F_{\beta}^{mean}~\uparrow$ & .391 & .649 & .547 & .562 & .634 & .642 & \underline{.674} & \textbf{.689}
  \\
  && $F_{\beta}^{W}~\uparrow$ & .319 & .615 & .478 & .535 & .593 & .601 & \underline{.637} & \textbf{.656}
  \\
  && $S_{\alpha}~\uparrow$ & .608 & .725 & .735 & .669 & .716 & .723 & \underline{.741} & \textbf{.755}
  \\
  && $E_{\phi}^{max}~\uparrow$ & .714 & .806 & .807 & .729 & .796 & .803 & \textbf{.827} & \underline{.822}
  \\
  && $E_{\phi}^{mean}~\uparrow$ & .565 & .802 & .683 & .719 & .777 & .784 & \textbf{.824} & \underline{.819}
  \\
  && $\mathcal{M}~\downarrow$ & .246 & .101 & .133 & .159 & .114 & .110 & \textbf{.084} & \underline{.085}
  \\

  \bottomrule
  \end{tabular}}
\end{table*}

\begin{figure*}[t!]
	\centering
	\begin{overpic}[width=0.95\textwidth]{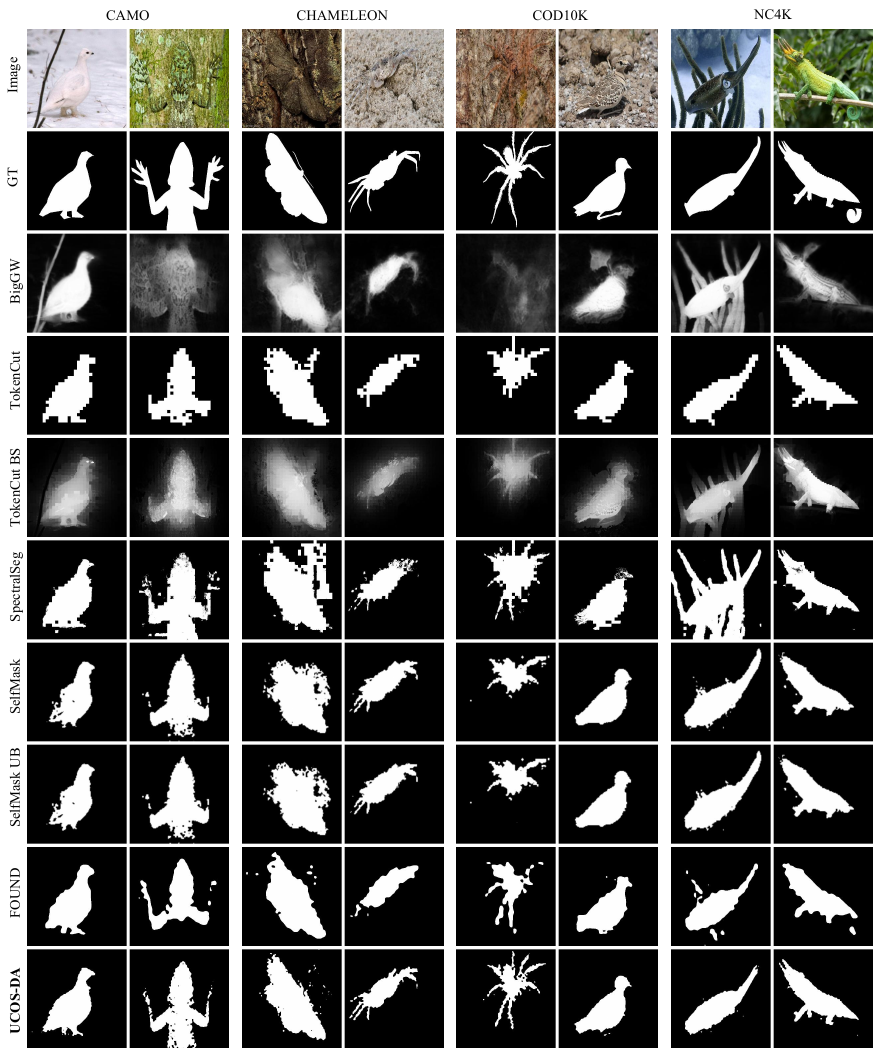}
    \end{overpic}
	\caption{Visual samples of our baseline model (\ourTask) and all competing models.}
    \label{fig:UCOSDA_visualization}
\end{figure*}

\subsection{Settings}

\noindent\textbf{Training DataSets.}
We randomly collect 300 images from the most commonly-used supervised COS training set \cite{fan2020camouflaged,zhang2023predictive}, which includes 4,040 images representing various camouflage-based scenes. We also randomly select 300 images from the most commonly-used salient object segmentation training set, \ie, DUTS-tr \cite{wang2017learning}. Thus, the training set for our \ourTask~consists of 600 images covering wide real-world daily scenes, while has its scale much smaller than the ones for fully-supervised image segmentation.

\noindent\textbf{Testing DataSets.}
To thoroughly analyze the performance of our new unsupervised baseline, we test our model and all benchmark models on six commonly-used testing sets, \ie, ECSSD \cite{shi2015hierarchical}, HKU-IS \cite{li2015visual}, CAMO \cite{le2019anabranch}, CHAMELEON \cite{Chameleon2018}, COD10K \cite{fan2020camouflaged} and NC4K \cite{yunqiu2021ranking}, which possess 1K, 4447, 250, 76, 2026 and 4121 images, respectively.

\noindent\textbf{Benchmark Models.}
To contribute the community a comprehensive benchmark towards unsupervised object segmentation, we collect most recent state-of-the-art fully unsupervised models, including BigGW \cite{voynov2021object}, TokenCut \cite{wang2022self}, SpectralSeg \cite{melas2022deep}, SelfMask \cite{Shin_2022_CVPR} and FOUND \cite{simeoni2022unsupervised}.

\noindent\textbf{Evaluation Metrics.}
We apply seven widely-used metrics to quantitatively evaluate all the benchmark models. The metrics include Accuracy ($Acc.$), mean Intersection over Union ($mIoU$), mean absolute error ($M$), F-measure \cite{Fmeasure} ($F_{\beta}$), weighted F-measure \cite{wfmeasure} ($F_{\beta}^{W}$), S-measure \cite{fan2017structure} ($S_\alpha$) and E-measure \cite{fan2018enhanced} ($E_\phi$). Notably, 

\noindent{$F_{\beta}$} computes both $Precision$ and $Recall$, formulated as:
\begin{equation}\label{fmeasure}
   F_{\beta} = \frac{(1+\beta^{2})Precision~Recall}{\beta^{2}Precision + Recall}, 
\end{equation}
with 
\begin{equation}
  Precision=\frac{\left|P\cap G\right|}{\left|P\right|}; Recall=\frac{\left|P\cap G\right|}{\left|G\right|},
\end{equation}
where $G$ is the ground truth and $P$ denotes a binarized predictions. Multiple $P$ are computed by assigning different integral thresholds $\tau$ ($\tau \in [0,255]$) to the predicted map. The $\beta^{\text{2}}$ is commonly set to 0.3 . 

\noindent{$S_{\alpha}$} evaluates the structural similarities between the prediction and the ground truth. The metric is defined as:
\begin{equation}\label{smeasure}
   S = \alpha S_{o} + (1 - \alpha) S_{r},
\end{equation}
where $S_{r}$ and $S_{o}$ denote the region-/object-based structure similarities, respectively. $\alpha \in [0,1]$ is empirically set as 0.5 to arrange equal weights to both region-level and object-level quantitative evaluation.

\noindent{$E_{\phi}$} is a cognitive vision-inspired metric evaluating both global and local similarities between two binary maps. The metric is defined as:
\begin{equation}\label{emeasure}
E_{\phi}=\frac{1}{WH}\sum_{i=1}^W\sum_{j=1}^H\phi\left(G(i,j), P(i,j)\right),
\end{equation} 
where $\phi$ represents the enhanced alignment matrix.

\subsection{Comparison with Unsupervised Methods}

\noindent\textbf{Zero-Shot Transfer.} As shown in Table \ref{tab:benchmark_UCOSDA_SOD} and Table \ref{tab:benchmark_UCOSDA_COD}, we benchmark all competing models on datasets for both camouflaged and salient object segmentation. As a result, our \ourTask~baseline model obtains overall superior performance on multiple testing sets. Please note that the benchmark results are all based on the codes and released checkpoints from each model's official project page. We also show some visual samples in Figure \ref{fig:UCOSDA_visualization}.

\begin{figure}[t!]
	\centering
	\begin{overpic}[width=0.48\textwidth]{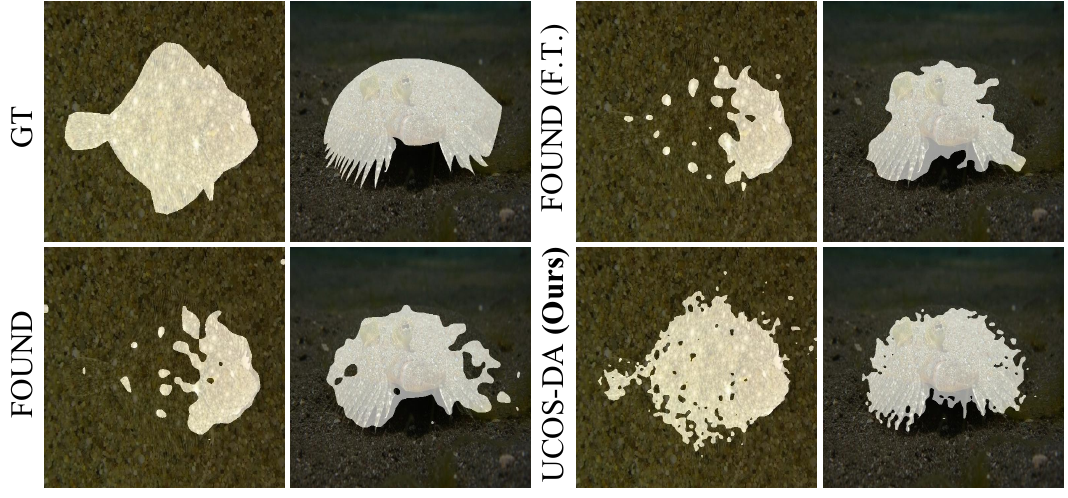}
    \end{overpic}
	\caption{Visual comparison of different linear-probe-based unsupervised image segmentation methods.}
    \label{fig:linear_probe}
\end{figure}

\noindent\textbf{Linear Probe via Adversarial training.} To analyze the effectiveness of our proposed \ourMod~module, we compare our results with the ones of FOUND \cite{simeoni2022unsupervised}, which also uses linear probe-based DINO fine-tuning strategy. As a result, we spot a slight performance drop of FOUND when fine-tuning its linear layer with COS training set (Table \ref{tab:benchmark_UCOSDA_Linear_FT}). We also show a visual example to further illustrate the phenomenon (Figure \ref{fig:linear_probe}). On the contrary, our method not only performs superior results on COS testing sets, but also acquires competitive results on salient object segmentation datasets, indicating the effectiveness and robustness of the proposed modules.

\begin{table}[t!]
  \centering
 \setlength\tabcolsep{3pt}
 \caption{
   Comparison of linear-probe strategies upon COS.
   }
   \label{tab:benchmark_UCOSDA_Linear_FT}
  \resizebox{0.48\textwidth}{!}{
  \begin{tabular}{rrrrrrr}
   \toprule
   \multirow{2}{*}{Method} & \multicolumn{3}{c}{COD10K} & \multicolumn{3}{c}{NC4K} 
   \\
   \cpartline{2-4} \cpartline{5-7}
   & mIoU~$\uparrow$ & Acc.~$\uparrow$ & $F_{\beta}^{max}~\uparrow$ 
   & mIoU~$\uparrow$ & Acc.~$\uparrow$ & $F_{\beta}^{max}~\uparrow$ 
   \\
   \cpartline{1-7}
   FOUND
   & 42.8 & \textbf{91.5} & 52.1 & 56.6 & \textbf{91.6} & 67.6
   \\ 

   FOUND (F.T.) 
   & \red{-4.0} & \red{-1.8} & \red{-4.8} & \red{-3.8} & \red{-1.5} & \red{-4.6}
   \\
   
   \textbf{Ours} 
   & \textbf{46.2} & 91.4 & \textbf{54.8} & \textbf{59.0} & 91.5 & \textbf{69.1}
   \\
  
   \bottomrule
  \end{tabular}}
\end{table}


\section{Conclusion and Future Work}
In this work, we investigate a new challenging image segmentation task, \ie, unsupervised camouflaged object segmentation. We firstly contribute a comprehensive benchmark study to show limited transferring ability of state-of-the-art unsupervised image segmentation models. We explore the co-existence of challenge and opportunity of a unique object-centric attribute, \ie, concealing coloration, and resort to prior-inspired adversarial domain adaptation to conduct the task. As a result, our new baseline model achieves overall superior scores based on multiple metrics and testing sets. 
Based on our study towards \ourTask, we find following issues that could be paid attention in the future researches.

\noindent\textbf{Attribute-based Domain Adaptation.} 
The concealing coloration attribute makes unsupervised COS an open issue in both societies of unsupervised domain adaptation and unsupervised image segmentation, in the context of current generic object datasets-based pre-trains. Future works may explore more towards specific domains where the objects own rarely-seen attributes, and investigate attribute-specific domain adaptation methods.  

\noindent\textbf{Generalizability of Self-Supervised Pre-trains.}
Our benchmark shows limited application of current self-supervised pre-trained models, which could inspire more studies towards generalizable pre-trains.
Investigating the transfer learning ability of self-supervised pre-trained models is essential, since it is expensive to train large models in each domain. Besides, exploring effective domain adaptation methods under challenging settings helps to advance the development of interpretable AI. We hope our preliminary work could inspire future researches towards more generalizable label-free segmentation and unsupervised domain adaptation methodologies.

\noindent\textbf{Other Learning Paradigms.} Besides \enquote{pre-training and fine-tuning}, future researches may explore unsupervised representation decomposition with attribute-sufficient real-world data, aiming to acquire both interpretability and generalizability.

{\small
\bibliographystyle{ieee_fullname}
\bibliography{egbib}
}

\end{document}